# *Fast Botnet Detection From Streaming Logs Using Online Lanczos Method*

Zheng Chen[2], Xinli Yu[3], Chi Zhang[4], Jin Zhang[1], Cui Lin[1],
Bo Song[2], Jianliang Gao[2], Xiaohua Hu[2], Wei-Shih Yang[3], Erjia Yan[2]
[1]CA Technologies, Inc.
[2] College of Computing & Informatics, Drexel University
[3]Department of Mathematics, Temple University
[4]Department of Computer Science, Maryland University at Baltimore County

*Abstract* — **Botnet, a group of coordinated bots, is becoming the main platform of malicious Internet activities like DDOS, click fraud, web scraping, spam/rumor distribution, etc. This paper focuses on design and experiment of a new approach for botnet detection from streaming web server logs, motivated by its wide applicability, real-time protection capability, ease of use and better security of sensitive data. Our algorithm is inspired by a Principal Component Analysis (PCA) to capture correlation in data, and we are first to recognize and adapt Lanczos method to improve the time complexity of PCA-based botnet detection from cubic to sub-cubic, which enables us to more accurately and sensitively detect botnets with sliding time windows rather than fixed time windows. We contribute a generalized online correlation matrix update formula, and a new termination condition for Lanczos iteration for our purpose based on error bound and non-decreasing eigenvalues of symmetric matrices. On our dataset of an ecommerce website logs, experiments show the time cost of Lanczos method with different time windows are consistently only 20% to 25% of PCA.**

## I. Introduction

A *bot* is a software application that runs automated scripts over the Internet [1] to perform malicious tasks like DOS attack, website statistics skew, click fraud, price/information scraping, spam/rumor distribution, etc. Traditional single bots usually execute their tasks at a rate much higher than average human users to achieve their goal within a time limit. Recent single-bot detection methods, like peak-finding [2], outlier detection [3], threat propagation [4], more or less use this property of single bots.

A *botnet*, as the name suggests, is a group of bots that work in a coordinated fashion. In contrast to single bots, a botnet, especially those large-scale botnets, they might request resources at a human-like speed, but altogether they place a heavy burden on the servers and collect large amount of information. Because bots in a botnet behave human-like, they are much harder to detect, and have become a key platform for many Internet attacks.

Log data have been commonly leveraged for both bot and botnet detection. Generally, we can say a *log* is a sequence of data entries order by timestamp, where each data entry carries several fields that record the properties of an activity at a specific time. The main ***objective*** of this paper is to develop an efficient botnet detection from large-scale streaming web server logs with host identifier, request identifier, and time stamp; for example, in this paper we will experiment on Apache HTTP access logs. A *host identifier* could be an IP addresses or a MAC addresses, or anything similar; a *request identifier* could be an URL or an IP address, or web API name, or anything similar; the concrete forms of those identifiers depend on the type of stream-in logs. For Apache HTTP access log, a host identifier could be an IP address, or a host name, and a request identifier is a URL pointing to some resource on the server. Although we focus on sever logs, however, the approach can be used to monitor any streaming log data with similar information for coordinated behavior.

The method developed in this paper is the key part of a larger bot/botnet detection system prototype overviewed in section II. The above objective is motivated by our research concerns, business goal and system requirements [5]. ***First*** is wider applicability. As far as we know, there still lacks researches on a generally applicable botnet detection method for web servers. Most recent botnet detection methods involve a particular type of non-log data. For example, [6] uses captcha test results to discover search engine bots, [7] takes advantage of an emulator to interact with the botnet, [8] is based on knowledge of protocol and DNS traffic, [9] needs to construct a user-user graph based on login activities. These methods typically intend for a special purpose and need additional efforts to collect and pre-process information not readily available on the server and sometimes need the modeler to understand advanced Internet structure. In contrast, our approach is quite "lightweight", which only relies on above-mentioned basic information present in various access/activity logs of computer systems/software that host network services. ***Secondly***, the time complexity. Users stream in their log data to our system; the system monitors the streaming log data and in turn provide real-time warning on potential malicious hosts. Such real-time feedback has requirement on method complexity as well as sensitivity to coordinated attack from botnet. Logs usually come in a large volume. Users of a medium-scale ecommerce website can generate 100,000 log entries in 30 minutes. More popular website could have a much bigger number. As far as we know, all previous papers mentioned here do not meet our requirement. For example, method like in [9] needs hundreds of computers to run hours to figure out the botnet. ***Thirdly***, we believe our approach will be more secure in the sense that users are not required to provide sensitive low-level hardware information. Our method also does not need to intercept and inspect Internet packets like [10] [11] [12], which could bring additional security concerns. Log is the only data we need, and our method even allows users to anonymize the identifiers before they stream in. ***Last not the least***, it is not possible for one method to detect all sorts of bots, and therefore modern industrial bot detection system is often an ensemble [13-15] integrating heterogenous methods to enhance detection capability. The simplicity and ease of use of our approach as discussed above: less data preprocessing, less requirement on specialized knowledge, less involvement with sensitive data, makes for better integration with other methods.

The method we propose to adapt is Lanczos iteration [16, 17]. It is a method in numerical linear algebra to estimate eigenvalues and eigenvectors, with rigorous theory on its error and convergence. Detailed discussion is in section III. This idea is inspired by Principal Component Analysis (PCA) that captures data correlation by computing eigenvalues eigenvectors of correlation matrix. The ***main contribution*** of this paper can be summarized as the following, to the extent of our knowledge: 1) we are the first to investigate PCA-based botnet detection from streaming logs; 2) we are the first to contribute an algorithm that updates correlation matrix for the most general case of sliding window in section III.B; 3) we are first to first to recognize and adapt Lanczos method for fast botnet detection, and we innovate on the termination condition in our setup using Theorem 1 and Theorem 2 that leads to early termination of each iteration; our experiments in section IV further shows its effectiveness.

## II. BACKGROUND

### A. Principal Component Analysis

Principal Component Analysis (PCA) is best known as a dimension reduction technique, is also a popular method in anomaly detection to detect outliers, for example, finding cyber anomalies [18, 19]; those data points not well-represented by the principal components are considered outliers or anomalies. It is also applied for the converse purpose to check if there is high-level coordination in the data, which has been successfully applied to the monitoring of industrial processes [20, 21]. PCA also has been combined with KL-divergence to detect botnet from search engine logs [22]; in that paper KL-divergence is used to filter users with usual "click distribution", but still full PCA with cubic complexity is applied to detect correlation is the user is deemed "unusual". A similar basic idea can also be found in [5].

Mathematically, PCA finds the eigenvalues and their eigenvectors of the covariance or correlation matrix s.t. the eigenvectors are orthonormal. Those orthonormal eigenvectors are then sorted by their eigenvalues and form rotated coordinate of the space and are called (first, second, …) principal components. In future discussion, we call the largest eigenvalue associated with the first principal component as the *principal weight*.

The major problem of PCA, or in particular the calculation of the principal weight from the correlation matrix, is its cubic time complexity. The technical purpose of this paper is to use Lanczos method to reduce this time complexity. We discuss more about this in section III.C.

### B. Bot Detection System Prototype

Our algorithm plays a key role in a bot detection prototype system, which has been filed for patent [5, 23, 24]. The general work flow of this system is illustrated in Figure 1, working side by side with a Markov chain-based behavior model [24]. The latter views bots from a different perspective and detects them by their strange activities on the sever. For example, a human visiting an ecommerce website usually first browse products, make several searches, log in, add product to carts, and check out. However, it is hard for bots to follow this routine. Such routine can be modelled by Markov chain transitions, and bot visit sequence will have low probability in this Markov chain. However, this behavior model can only detect single bots or botnet-bots individually, lacking the ability to discover a botnet as a whole. As mentioned earlier, we favor a lightweight botnet detection algorithm, so it can be more easily integrated with the workflow.

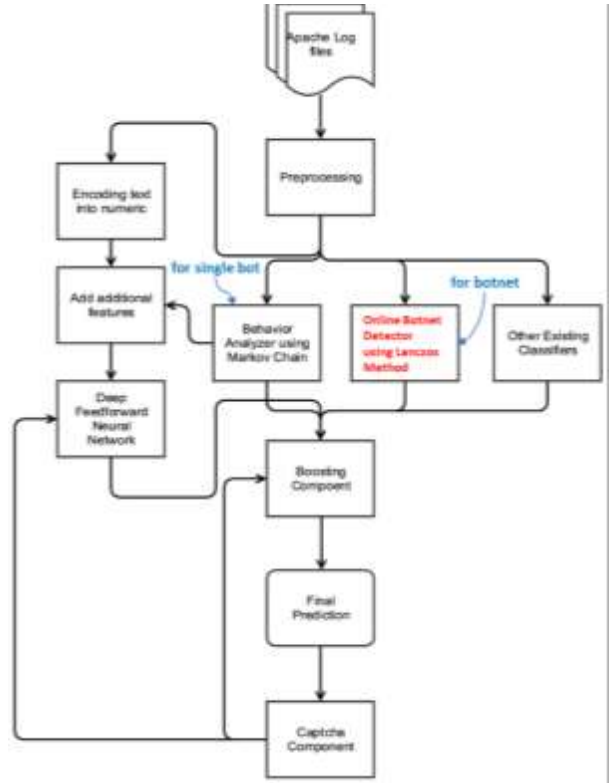

Figure 1 The workflow of our bot/botnet detection prototype system.

## III. PROBLEM & APPROACH

### A. Problem Formulation

For botnet detection of streaming log data, PCA provides a good start point, and a straightforward solution could be as the following: 1) split the streaming logs into time windows according to a specified interval; 2) for each time window, convert the log entries within a specified time interval to a host-request matrix **X** with the integer value at $i$th row and $j$th column, denoted by $\mathbf{X}(i,j)$, represents the number of times host $j$ makes request $i$; 3) run PCA on each time window, and PCA can be applied on **X** to check if the principal weight exceeds certain threshold. Above procedure can detect either single bots with a large volume of traffic, or a botnet where each bot might not make many requests but they correlate with each other and altogether still produce a high volume of traffic. An example is illustrated in Figure 2.

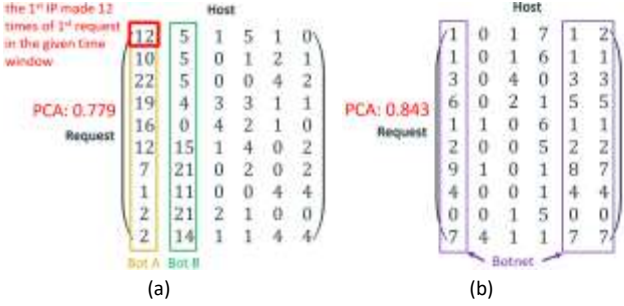

Figure 2 (a) Two single bots, each of which has clearly higher visit rates that non-bot hosts. Although their visit counts are not correlated, they dominate non-bot visits, leading to a high principal weight of 0.779. (b) A botnet with correlated visit counts, with a high principal weight of 0.843.

Nonetheless, questions arise in the adaptation of PCA to our application. ***First***, the time complexity of full PCA used in [22] is cubic, which easily breaks down when a large number of logs come in. In our application, we are probably interested in only the largest eigenvalue and its eigenvector. It is obviously not necessary for a full PCA. ***Secondly***, division into time windows seems not suitable for monitor of streaming data. We are facing an *accuracy-sensitivity dilemma*: a small window will weaken the algorithm's ability to find bots, since fewer data might not provide sufficient evidence for PCA; a larger window will slow down the algorithm's sensitivity, e.g. in a massive attack the sever might already be brought down before the algorithm starts to analyze the last 30-min window. There is no universal criterion for good window length, and it is difficult to conduct experiments for different server logs. A more professional practice is *sliding window*, as in industrial process monitor [21, 25, 26], but this makes computation even more intense. In process control, the "features" as columns of data matrix is usually at most hundreds of columns, while in our application, the "features" are tens of thousands of host identifiers.

From above discussion, reducing time complexity is the top priority if we desire to use sliding window. Our ***research problem*** can thus be summarized as constructing a fast algorithm suitable for using sliding window to monitor large-scale streaming logs for potential bots/botnets.

### B. Correlation Matrix Update

Consider the request-host matrix introduced in previous section. We evaluate the principal weight from the correlation matrix between hosts. After each window slide, this matrix will change, so does the correlation matrix. Our first problem is how to update the correlation matrix after every time window slide before updating the principal weight. The time complexity for computing correlation matrix of a $n \times m$ request-host matrix is $O(nm^2)$, so it is highly uneconomic and undesirable to re-compute entire correlation matrix. [21] calculated the updates for adding new rows and deleting old rows. Our case is more complicated: after every window slide, some rows and columns might be removed, some new rows and columns might be appended, and other rows and columns might have value change.

Let $\mathbf{X}_t^o \in \mathbb{R}^{n_t \times m_t}$ be the request-host matrix before the $t$th window slide, $t \in \mathbb{N}^+$. Then the mean of each column is given in the vector

$$\mathbf{b}_t = \frac{1}{n_t}(\mathbf{X}_t^o)^\mathrm{T} \mathbf{1}_{n_t} \quad (1)$$

where $\mathbf{1}_{n_t} = [1, \dots, 1]^\mathrm{T} \in \mathbb{R}^{n_t}$ is a column vector of length $n_t$ with all components being 1. Then we call the following as the *centralize request-host matrix*, because each of its column has zero mean

$$\mathbf{X}_t = \mathbf{X}_t^o - \mathbf{1}_{n_t}\mathbf{b}_t^\mathrm{T} \quad (2)$$

where $\mathbf{\Sigma}_t = \mathrm{diag}(\sigma_{t,1}, \dots, \sigma_{t,m_t})$ is a diagonal matrix with the $j$th diagonal element $\sigma_{t,j}$ being the standard deviation of the $j$th column of $\mathbf{X}_t^o$. The correlation matrix for $\mathbf{X}_t^o$ is thus

$$\mathbf{R}_t = \frac{1}{n_t - 1} \mathbf{\Sigma}_t^{-1} \mathbf{X}_t^\mathrm{T} \mathbf{X}_t \mathbf{\Sigma}_t^{-1} \quad (3)$$

Our goal is to find the update formula for $\mathbf{R}_{t+1}$ using information from $\mathbf{X}_t^o, \mathbf{b}_t, \mathbf{X}_t$ and $\mathbf{R}_t$. For simplicity, we can assume that the hosts are the same before and after the window slide, without loss of generality, i.e. column changes can be disregarded when calculating the updates. This is because, if there are any added columns or removed columns caused by window slide, we can simply add corresponding columns (and rows if necessary) into $\mathbf{b}_t, \mathbf{X}_t, \mathbf{R}_t$, and use the modified ones for our future inferences.

Also note correlation is not dependent on row order, therefore rows of $\mathbf{X}_t$ can be arbitrarily arranged for our convenience. Suppose after the $t$th slide,

1) $n_{t+1}^+$ new rows $\mathbf{X}_{n_{t+1}^+}^o$ are appended to the bottom of $\mathbf{X}_t^o$, each row representing a new request-host vector not currently in $\mathbf{X}_t^o$;

2) $n_{t+1}^-$ rows $\mathbf{X}_{n_{t+1}^-}^o$ on top of $\mathbf{X}_t^o$ are removed, meaning those requests disappear after the window slide;

3) other rows have value change denoted by matrix $\mathbf{X}_{c_{t+1}}^o$, where only the top $c_{t+1}$ of $\mathbf{X}_{c_{t+1}}^o$ are non-zero.

Since every window slide is small and of fixed distance, thus $n_{t+1}^+ + n_{t+1}^- + c_{t+1}$ is usually small and can be treated as a constant. In future discussion, we also write $\mathbf{X}_t^o = \begin{bmatrix} \mathbf{X}_{n_{t+1}^-}^o \\ \mathbf{X}_{n_t - n_{t+1}^-}^o \end{bmatrix}$ where $\mathbf{X}_{n_t - n_{t+1}^-}^o$ is the last $n_t - n_{t+1}^-$ rows of $\mathbf{X}_t^o$; likewise, $\mathbf{X}_t = \begin{bmatrix} \mathbf{X}_{n_{t+1}^-} \\ \mathbf{X}_{n_t - n_{t+1}^-} \end{bmatrix}$ where $\mathbf{X}_{n_{t+1}^-}$ and $\mathbf{X}_{n_t - n_{t+1}^-}$ are centralized data of $\mathbf{X}_{n_{t+1}^-}^o$ and $\mathbf{X}_{n_t - n_{t+1}^-}^o$ respectively like in (2). We have the following relation for $\mathbf{X}_{n_{t+1}^+}^o, \mathbf{X}_{n_{t+1}^-}^o$, and $\mathbf{X}_{c_{t+1}}^o$, where $\mathbf{0}_{n_{t+1}^- \times m_t}$ denotes a $n_{t+1}^- \times m_t$ zero matrix, and $\mathbf{X}_{t+1}^{*o}$ is equivalent to $\mathbf{X}_{t+1}^o$ except for the remove rows are replaced by zeros.

$$\begin{bmatrix} \mathbf{0}_{n_{t+1}^- \times m_t} \\ \mathbf{X}_{t+1}^o \end{bmatrix} = \begin{bmatrix} \mathbf{X}_t^o - \begin{bmatrix} \mathbf{X}_{n_{t+1}^-}^o \\ \mathbf{X}_{c_{t+1}}^o \end{bmatrix} \\ \mathbf{X}_{n_{t+1}^+}^o \end{bmatrix} := \mathbf{X}_{t+1}^{*o} \quad (4)$$

The update of mean-value is now given by

$$(n_t + n_{t+1}^+ - n_{t+1}^-)\mathbf{b}_{t+1} = n_t\mathbf{b}_t + \left(\mathbf{X}_{n_{t+1}^+}^o\right)^T \mathbf{1}_{n_{t+1}^+} - \left(\mathbf{X}_{n_{t+1}^-}^o\right)^T \mathbf{1}_{n_{t+1}^-} + \left(\mathbf{X}_{c_{t+1}}^o\right)^T \mathbf{1}_{c_{t+1}} \quad (5)$$

Let $\Delta \mathbf{b}_{t+1}^T = \mathbf{b}_{t+1}^T - \mathbf{b}_t^T$, and note $\mathbf{X}_{n_{t+1}^+} = \mathbf{X}_{n_{t+1}^+}^o - \mathbf{1}_{n_{t+1}^+}\mathbf{b}_{t+1}^T$ is data centralization like in (2), we then have

$$\mathbf{X}_{t+1}^* = \begin{bmatrix} \mathbf{0}_{n_{t+1}^- \times m_t} \\ \mathbf{X}_{t+1}^o \end{bmatrix} - \begin{bmatrix} \mathbf{0}_{n_{t+1}^- \times m_t} \\ \mathbf{1}_{n_{t+1}}\mathbf{b}_{t+1}^T \end{bmatrix}$$

$$= \begin{bmatrix} \mathbf{X}_t^o - \mathbf{1}_{n_t}\mathbf{b}_t^T + \mathbf{1}_{n_t}\mathbf{b}_t^T - \begin{bmatrix} \mathbf{X}_{n_{t+1}^-}^o \\ \mathbf{X}_{c_{t+1}}^o \end{bmatrix} \\ \mathbf{X}_{n_{t+1}^+}^o \end{bmatrix} - \begin{bmatrix} \mathbf{0}_{n_{t+1}^- \times m_t} \\ \mathbf{1}_{n_{t+1}}\mathbf{b}_{t+1}^T \end{bmatrix}$$

$$= \begin{bmatrix} \mathbf{X}_t^o - \mathbf{1}_{n_t}\mathbf{b}_t^T + \mathbf{1}_{n_t}\mathbf{b}_t^T - \begin{bmatrix} \mathbf{X}_{n_{t+1}^-}^o \\ \mathbf{X}_{c_{t+1}}^o \end{bmatrix} - \begin{bmatrix} \mathbf{0}_{n_{t+1}^- \times m_t} \\ \mathbf{1}_{n_t-n_{t+1}^-}\mathbf{b}_{t+1}^T \end{bmatrix} \\ \mathbf{X}_{n_{t+1}^+}^o - \mathbf{1}_{n_{t+1}^+}\mathbf{b}_{t+1}^T \end{bmatrix}$$

$$= \begin{bmatrix} \mathbf{X}_t + \left(\mathbf{1}_{n_t}\mathbf{b}_t^T - \begin{bmatrix} \mathbf{X}_{n_{t+1}^-}^o \\ \mathbf{X}_{c_{t+1}}^o \end{bmatrix} - \begin{bmatrix} \mathbf{0}_{n_{t+1}^- \times m_t} \\ \mathbf{1}_{n_t-n_{t+1}^-}\mathbf{b}_{t+1}^T \end{bmatrix}\right) \\ \mathbf{X}_{n_{t+1}^+} \end{bmatrix} \quad (6)$$

$$= \begin{bmatrix} \mathbf{X}_t - \begin{bmatrix} \mathbf{X}_{n_{t+1}^-}^o - \mathbf{1}_{n_{t+1}^-}\mathbf{b}_t^T \\ \mathbf{X}_{c_{t+1}}^o + \mathbf{1}_{n_t-n_{t+1}^-}(\mathbf{b}_{t+1}^T - \mathbf{b}_t^T) \end{bmatrix} \\ \mathbf{X}_{n_{t+1}^+} \end{bmatrix}$$

$$= \begin{bmatrix} \begin{bmatrix} \mathbf{0}_{n_{t+1}^- \times m_t} \\ \overbrace{\mathbf{X}_{n_t-n_{t+1}^-} - \mathbf{X}_{c_{t+1}}^o - \mathbf{1}_{n_t-n_{t+1}^-}\Delta\mathbf{b}_{t+1}^T}^{o(m_t(n_t-n_{t+1}^-))} \end{bmatrix} \\ \underbrace{\mathbf{X}_{n_{t+1}^+}}_{o(m_t n_{t+1}^+)} \end{bmatrix}$$

where we note the top $n_{t+1}^-$ rows of $\mathbf{X}_{t+1}^*$ must all be zero, and thus the time complexity of $\mathbf{X}_{t+1}^*$ is $O(m_t(n_t - n_{t+1}^- + n_{t+1}^+)) = O(m_t n_{t+1})$. We now start calculating the update for the standard deviations in $\mathbf{\Sigma}_{t+1}$.

$$\sigma_{t+1,j}^2 = \frac{\left\| \begin{bmatrix} \mathbf{X}_t^o(:,j) - \begin{bmatrix} \mathbf{X}_{n_{t+1}^-}^o(:,j) \\ \mathbf{X}_{c_{t+1}}^o(:,j) \end{bmatrix} \\ \mathbf{X}_{n_{t+1}^+}^o(:,j) \end{bmatrix} - \begin{bmatrix} \mathbf{0}_{n_{t+1}^-} \\ \mathbf{b}_{t+1}(j)\mathbf{1}_{n_{t+1}} \end{bmatrix} \right\|^2}{n_{t+1}-1}$$

$$= \frac{\left\| \begin{bmatrix} \mathbf{X}_t^o(:,j) - \mathbf{1}_{n_t}\mathbf{b}_t^T(j) + \mathbf{1}_{n_t}\mathbf{b}_t^T(j) - \begin{bmatrix} \mathbf{X}_{n_{t+1}^-}^o(:,j) \\ \mathbf{X}_{c_{t+1}}^o(:,j) \end{bmatrix} - \begin{bmatrix} \mathbf{0}_{n_{t+1}^-} \\ \mathbf{b}_{t+1}(j)\mathbf{1}_{n_t-n_{t+1}^-} \end{bmatrix} \\ \mathbf{X}_{n_{t+1}^+}^o(:,j) - \mathbf{b}_{t+1}(j)\mathbf{1}_{n_{t+1}^+} \end{bmatrix} \right\|^2}{n_{t+1}-1}$$

(7)

We can simplify (7) piece by piece. First expand,

$$\left\| \mathbf{X}_t^o(:,j) - \mathbf{b}_t(j)\mathbf{1}_{n_t} + \mathbf{b}_t(j)\mathbf{1}_{n_t} - \begin{bmatrix} \mathbf{X}_{n_{t+1}^-}^o(:,j) \\ \mathbf{X}_{c_{t+1}}^o(:,j) \end{bmatrix} - \begin{bmatrix} \mathbf{0}_{n_{t+1}^-} \\ \mathbf{b}_{t+1}(j)\mathbf{1}_{n_t-n_{t+1}^-} \end{bmatrix} \right\|$$

$$= \left\| \mathbf{X}_t^o(:,j) - \mathbf{b}_t(j)\mathbf{1}_{n_t} \right\|^2 + \left\| \mathbf{b}_t(j)\mathbf{1}_{n_t} - \begin{bmatrix} \mathbf{X}_{n_{t+1}^-}^o(:,j) \\ \mathbf{X}_{c_{t+1}}^o(:,j) \end{bmatrix} - \begin{bmatrix} \mathbf{0}_{n_{t+1}^-} \\ \mathbf{b}_{t+1}(j)\mathbf{1}_{n_t-n_{t+1}^-} \end{bmatrix} \right\|^2$$

$$+ 2\left[ \mathbf{b}_t(j)\mathbf{1}_{n_t} - \begin{bmatrix} \mathbf{X}_{n_{t+1}^-}^o(:,j) \\ \mathbf{X}_{c_{t+1}}^o(:,j) \end{bmatrix} - \begin{bmatrix} \mathbf{0}_{n_{t+1}^-} \\ \mathbf{b}_{t+1}(j)\mathbf{1}_{n_t-n_{t+1}^-} \end{bmatrix} \right]^T [\mathbf{X}_t^o(:,j) - \mathbf{b}_t(j)\mathbf{1}_{n_t}]$$

(8)

where

$$\left\| \mathbf{X}_t^o(:,j) - \mathbf{b}_t(j)\mathbf{1}_{n_t} \right\|^2 = (n_t - 1)\sigma_{t,j}^2 \quad (9)$$

$$\mathbf{b}_t(j)\mathbf{1}_{n_t} - \begin{bmatrix} \mathbf{X}_{n_{t+1}^-}^o(:,j) \\ \mathbf{X}_{c_{t+1}}^o(:,j) \end{bmatrix} - \begin{bmatrix} \mathbf{0}_{n_{t+1}^-} \\ \mathbf{b}_{t+1}(j)\mathbf{1}_{n_t-n_{t+1}^-} \end{bmatrix}$$
$$= \begin{bmatrix} \mathbf{b}_t(j)\mathbf{1}_{n_t} - \mathbf{X}_{n_{t+1}^-}^o(:,j) \\ -\Delta\mathbf{b}_{t+1}(j)\mathbf{1}_{n_t-n_{t+1}^-} - \mathbf{X}_{c_{t+1}}^o(:,j) \end{bmatrix} \quad (10)$$

$$\mathbf{1}_{n_t}^T \mathbf{X}_t^o(:,j) - \mathbf{1}_{n_t}^T \mathbf{b}_t(j)\mathbf{1}_{n_t} = 0$$
$$\Rightarrow [\mathbf{b}_t(j)\mathbf{1}_{n_t}]^T [\mathbf{X}_t^o(:,j) - \mathbf{b}_t(j)\mathbf{1}_{n_t}] = 0 \quad (11)$$

Using (11), we have the following important simplification.

$$\begin{bmatrix} \mathbf{0}_{n_{t+1}^-} \\ \mathbf{b}_{t+1}(j)\mathbf{1}_{n_t-n_{t+1}^-} \end{bmatrix}^T [\mathbf{X}_t^o(:,j) - \mathbf{b}_t(j)\mathbf{1}_{n_t}]$$

$$= \left( \mathbf{b}_{t+1}(j)\mathbf{1}_{n_t}^T - \begin{bmatrix} \mathbf{b}_{t+1}(j)\mathbf{1}_{n_{t+1}^-} \\ \mathbf{0}_{n_t-n_{t+1}^-} \end{bmatrix}^T \right) [\mathbf{X}_t^o(:,j) - \mathbf{b}_t(j)\mathbf{1}_{n_t}] \quad (12)$$

$$= -\begin{bmatrix} \mathbf{b}_{t+1}(j)\mathbf{1}_{n_{t+1}^-} \\ \mathbf{0}_{n_t-n_{t+1}^-} \end{bmatrix}^T \mathbf{X}_t(:,j)$$

Plug (9)~(12) back to (8) and then (7), we have

$$(n_{t+1} - 1)\sigma_{t+1,j}^2$$

$$= (n_t - 1)\sigma_{t,j}^2 + \left\| \begin{bmatrix} \overbrace{\mathbf{b}_t(j)\mathbf{1}_{n_t} - \mathbf{X}_{n_{t+1}^-}^o(:,j)}^{o(n_{t+1}^-)} \\ \overbrace{\Delta\mathbf{b}_{t+1}(j)\mathbf{1}_{n_t-n_{t+1}^-} + \mathbf{X}_{c_{t+1}}^o(:,j)}^{o(c_{t+1})} \end{bmatrix} \right\|^2$$

$$- 2\overbrace{\begin{bmatrix} \mathbf{X}_{n_{t+1}^-}^o(:,j) + \mathbf{b}_{t+1}(j)\mathbf{1}_{n_{t+1}^-} \\ \mathbf{X}_{c_{t+1}}^o(:,j) \end{bmatrix}^T}^{o(n_{t+1}^- + c_{t+1})} \mathbf{X}_t(:,j) + \left\| \overbrace{\mathbf{X}_{n_{t+1}^+}^o(:,j) - \mathbf{b}_{t+1}(j)\mathbf{1}_{n_{t+1}^+}}^{o(n_{t+1}^+)} \right\|^2$$

(13)

Recall the top $c_{t+1}$ rows of $\mathbf{X}_{c_{t+1}}^o$ are non-zero, thus from (13) the time complexity for updating each standard deviation is $O(n_{t+1}^- + c_{t+1} + n_{t+1}^+)$, linear to the number of rows that are affected by the window slide. The total complexity for updating all standard deviations is $O(m_t(n_{t+1}^- + c_{t+1} + n_{t+1}^+))$. At last we update the correlation matrix. Using the fourth identity of (6) and a technique like (12), we have

$$(n_{t+1} - 1)\mathbf{R}_{t+1} = (\mathbf{X}_{t+1}^* \mathbf{\Sigma}_{t+1}^{-1})^T (\mathbf{X}_{t+1}^* \mathbf{\Sigma}_{t+1}^{-1})$$

$$= \left( \mathbf{X}_t \mathbf{\Sigma}_{t+1}^{-1} + \begin{bmatrix} \left(\mathbf{1}_{n_t}\mathbf{b}_t^T - \begin{bmatrix} \mathbf{X}_{n_{t+1}^-}^o \\ \mathbf{X}_{c_{t+1}}^o \end{bmatrix} - \begin{bmatrix} \mathbf{0}_{n_{t+1}^- \times m_t} \\ \mathbf{1}_{n_t-n_{t+1}^-}\mathbf{b}_{t+1}^T \end{bmatrix} \right) \mathbf{\Sigma}_{t+1}^{-1} \\ \mathbf{X}_{n_{t+1}^+}\mathbf{\Sigma}_{t+1}^{-1} \end{bmatrix} \right)^T$$

$$\times \left( \mathbf{X}_t \mathbf{\Sigma}_{t+1}^{-1} + \begin{bmatrix} \left(\mathbf{1}_{n_t}\mathbf{b}_t^T - \begin{bmatrix} \mathbf{X}_{n_{t+1}^-}^o \\ \mathbf{X}_{c_{t+1}}^o \end{bmatrix} - \begin{bmatrix} \mathbf{0}_{n_{t+1}^- \times m_t} \\ \mathbf{1}_{n_t-n_{t+1}^-}\mathbf{b}_{t+1}^T \end{bmatrix} \right) \mathbf{\Sigma}_{t+1}^{-1} \\ \mathbf{X}_{n_{t+1}^+}\mathbf{\Sigma}_{t+1}^{-1} \end{bmatrix} \right) \quad (14)$$

$$= \mathbf{\Sigma}_{t+1}^{-1}\mathbf{X}_t^T\mathbf{X}_t\mathbf{\Sigma}_{t+1}^{-1} + \mathbf{\Sigma}_{t+1}^{-1}\mathbf{Y}_{t+1}\mathbf{\Sigma}_{t+1}^{-1}$$

$$= \overbrace{(n_t - 1)\mathbf{\Sigma}_{t+1}^{-1}\mathbf{\Sigma}_t\mathbf{R}_t\mathbf{\Sigma}_t\mathbf{\Sigma}_{t+1}^{-1}}^{o(m_t^2)} + \overbrace{\mathbf{\Sigma}_{t+1}^{-1}\mathbf{Y}_{t+1}\mathbf{\Sigma}_{t+1}^{-1}}^{o(m_t^2)}$$

where

$$\mathbf{Y}_{t+1} = 2\mathbf{X}_t^T \left( \mathbf{1}_{n_t}\mathbf{b}_t^T - \begin{bmatrix} \mathbf{X}_{n_{t+1}^-}^o \\ \mathbf{X}_{c_{t+1}}^o \end{bmatrix} - \begin{bmatrix} \mathbf{0}_{n_{t+1}^- \times m_t} \\ \mathbf{1}_{n_t-n_{t+1}^-}\mathbf{b}_{t+1}^T \end{bmatrix} \right)$$

$$+ \left( \mathbf{1}_{n_t}\mathbf{b}_t^T - \begin{bmatrix} \mathbf{X}_{n_{t+1}^-}^o \\ \mathbf{X}_{c_{t+1}}^o \end{bmatrix} - \begin{bmatrix} \mathbf{0}_{n_{t+1}^- \times m_t} \\ \mathbf{1}_{n_t-n_{t+1}^-}\mathbf{b}_{t+1}^T \end{bmatrix} \right)^T \left( \mathbf{1}_{n_t}\mathbf{b}_t^T - \begin{bmatrix} \mathbf{X}_{n_{t+1}^-}^o \\ \mathbf{X}_{c_{t+1}}^o \end{bmatrix} - \begin{bmatrix} \mathbf{0}_{n_{t+1}^- \times m_t} \\ \mathbf{1}_{n_t-n_{t+1}^-}\mathbf{b}_{t+1}^T \end{bmatrix} \right)$$

$$+ \mathbf{X}_{n_{t+1}^+}^T \mathbf{X}_{n_{t+1}^+}$$

(15)

We then use the following identities multiple times

$$\mathbf{X}_t^T \mathbf{1}_{n_t} = \mathbf{0} \Rightarrow \mathbf{X}_t^T \mathbf{X}_t^o = \mathbf{X}_t^T(\mathbf{X}_t^o - \mathbf{1}_{n_t}\mathbf{b}_t^T + \mathbf{1}_{n_t}\mathbf{b}_t^T) \\
= \mathbf{X}_t^T \mathbf{X}_t + \mathbf{X}_t^T \mathbf{1}_{n_t}\mathbf{b}_t^T = \mathbf{X}_t^T \mathbf{X}_t \tag{16}$$

$$\begin{bmatrix} \mathbf{0}_{n_{t+1}^- \times m_t} \\ \mathbf{1}_{n_t - n_{t+1}^-}\mathbf{b}_{t+1}^T \end{bmatrix} = \mathbf{1}_{n_t}\mathbf{b}_{t+1}^T - \begin{bmatrix} \mathbf{1}_{n_{t+1}^-}\mathbf{b}_{t+1}^T \\ \mathbf{0}_{(n_t - n_{t+1}^-) \times m_t} \end{bmatrix} \tag{17}$$

$$\Delta \mathbf{b}_{t+1} \mathbf{1}_{n_t}^T \mathbf{1}_{n_t} \Delta \mathbf{b}_{t+1}^T = n_t \Delta \mathbf{b}_{t+1} \Delta \mathbf{b}_{t+1}^T \tag{18}$$

to arrive at

$$2\mathbf{X}_t^T \left( \begin{bmatrix} \mathbf{X}_{n_{t+1}^-}^o \\ \mathbf{X}_{c_{t+1}}^o \end{bmatrix} + \begin{bmatrix} \mathbf{0}_{n_{t+1}^- \times m_t} \\ \mathbf{1}_{n_t - n_{t+1}^-}\mathbf{b}_{t+1}^T \end{bmatrix} \right) \\
= 2\mathbf{X}_t^T \left( \mathbf{1}_{n_t}\mathbf{b}_{t+1}^T + \begin{bmatrix} \mathbf{X}_{n_{t+1}^-}^o \\ \mathbf{X}_{c_{t+1}}^o \end{bmatrix} - \begin{bmatrix} \mathbf{1}_{n_{t+1}^-}\mathbf{b}_{t+1}^T \\ \mathbf{0}_{(n_t - n_{t+1}^-) \times m_t} \end{bmatrix} \right) \\
= 2\mathbf{X}_t^T \begin{bmatrix} \mathbf{X}_{n_{t+1}^-}^o - \mathbf{1}_{n_{t+1}^-}\mathbf{b}_{t+1}^T \\ \mathbf{X}_{c_{t+1}}^o \end{bmatrix} \\
= 2(\mathbf{X}_{n_{t+1}^-}^T \mathbf{X}_{n_{t+1}^-} + \mathbf{X}_{n_t - n_{t+1}^-}^T \mathbf{X}_{c_{t+1}}^o) \tag{19}$$

and

$$\left( \mathbf{1}_{n_t}\mathbf{b}_t^T - \begin{bmatrix} \mathbf{X}_{n_{t+1}^-}^o \\ \mathbf{X}_{c_{t+1}}^o \end{bmatrix} - \begin{bmatrix} \mathbf{0}_{n_{t+1}^- \times m_t} \\ \mathbf{1}_{n_t - n_{t+1}^-}\mathbf{b}_{t+1}^T \end{bmatrix} \right)^T \left( \mathbf{1}_{n_t}\mathbf{b}_t^T - \begin{bmatrix} \mathbf{X}_{n_{t+1}^-}^o \\ \mathbf{X}_{c_{t+1}}^o \end{bmatrix} - \begin{bmatrix} \mathbf{0}_{n_{t+1}^- \times m_t} \\ \mathbf{1}_{n_t - n_{t+1}^-}\mathbf{b}_{t+1}^T \end{bmatrix} \right) \\
= \left( \mathbf{1}_{n_t}\Delta\mathbf{b}_{t+1}^T + \begin{bmatrix} \mathbf{X}_{n_{t+1}^-}^o - \mathbf{1}_{n_{t+1}^-}\mathbf{b}_{t+1}^T \\ \mathbf{X}_{c_{t+1}}^o \end{bmatrix} \right)^T \left( \mathbf{1}_{n_t}\Delta\mathbf{b}_{t+1}^T + \begin{bmatrix} \mathbf{X}_{n_{t+1}^-}^o - \mathbf{1}_{n_{t+1}^-}\mathbf{b}_{t+1}^T \\ \mathbf{X}_{c_{t+1}}^o \end{bmatrix} \right) \\
= n_t \Delta\mathbf{b}_{t+1}\Delta\mathbf{b}_{t+1}^T + 2\Delta\mathbf{b}_{t+1}\mathbf{1}_{n_t}^T \begin{bmatrix} \mathbf{X}_{n_{t+1}^-}^o - \mathbf{1}_{n_{t+1}^-}\mathbf{b}_{t+1}^T \\ \mathbf{X}_{c_{t+1}}^o \end{bmatrix} \Sigma_{t+1}^{-1} \\
+ \begin{bmatrix} \mathbf{X}_{n_{t+1}^-}^o - \mathbf{1}_{n_{t+1}^-}\mathbf{b}_{t+1}^T \\ \mathbf{X}_{c_{t+1}}^o \end{bmatrix}^T \begin{bmatrix} \mathbf{X}_{n_{t+1}^-}^o - \mathbf{1}_{n_{t+1}^-}\mathbf{b}_{t+1}^T \\ \mathbf{X}_{c_{t+1}}^o \end{bmatrix} \tag{20}$$

Continue the simplification (sketch calculation, some steps are long and hence omitted, but they are not hard to verify).

$$\Delta\mathbf{b}_{t+1}\mathbf{1}_{n_t}^T \begin{bmatrix} \mathbf{X}_{n_{t+1}^-}^o - \mathbf{1}_{n_{t+1}^-}\mathbf{b}_{t+1}^T \\ \mathbf{X}_{c_{t+1}}^o \end{bmatrix} \\
= \Delta\mathbf{b}_{t+1}(\mathbf{1}_{n_{t+1}^-}^T \mathbf{X}_{n_{t+1}^-}^o - \mathbf{1}_{n_{t+1}^-}^T \mathbf{1}_{n_{t+1}^-}\mathbf{b}_{t+1}^T + \mathbf{1}_{n_t - n_{t+1}^-}^T \mathbf{X}_{c_{t+1}}^o) \\
= \Delta\mathbf{b}_{t+1}(\mathbf{1}_{n_{t+1}^-}^T \mathbf{X}_{n_{t+1}^-}^o + \mathbf{1}_{n_t - n_{t+1}^-}^T \mathbf{X}_{c_{t+1}}^o - n_{t+1}^- \mathbf{b}_{t+1}^T) \tag{21}$$

$$\begin{bmatrix} \mathbf{X}_{n_{t+1}^-}^o - \mathbf{1}_{n_{t+1}^-}\mathbf{b}_{t+1}^T \\ \mathbf{X}_{c_{t+1}}^o \end{bmatrix}^T \begin{bmatrix} \mathbf{X}_{n_{t+1}^-}^o - \mathbf{1}_{n_{t+1}^-}\mathbf{b}_{t+1}^T \\ \mathbf{X}_{c_{t+1}}^o \end{bmatrix} \\
= (\mathbf{X}_{n_{t+1}^-}^o - \mathbf{1}_{n_{t+1}^-}\mathbf{b}_{t+1}^T)^T(\mathbf{X}_{n_{t+1}^-}^o - \mathbf{1}_{n_{t+1}^-}\mathbf{b}_{t+1}^T) + (\mathbf{X}_{c_{t+1}}^o)^T(\mathbf{X}_{c_{t+1}}^o) \\
= \mathbf{X}_{n_{t+1}^-}^T \mathbf{X}_{n_{t+1}^-} + (\mathbf{X}_{c_{t+1}}^o)^T(\mathbf{X}_{c_{t+1}}^o) + n_{t+1}^- \Delta\mathbf{b}_{t+1}\Delta\mathbf{b}_{t+1}^T \tag{22}$$

Finally, plug (21) and (22) back to (20), and then plug (19) and (20) back to (15), and after certain rearrangement, we will have

$$\mathbf{Y}_{t+1} = \overbrace{(n_t - n_{t+1}^-)\Delta\mathbf{b}_{t+1}\Delta\mathbf{b}_{t+1}^T}^{O(m_t^2)} + \overbrace{\mathbf{X}_{n_{t+1}^+}^T \mathbf{X}_{n_{t+1}^+}}^{O(m_t^2 n_{t+1}^+)} \\
+ \overbrace{2\Delta\mathbf{b}_{t+1}\mathbf{1}_{n_{t+1}^-}^T \mathbf{X}_{n_{t+1}^-}^o - \mathbf{X}_{n_{t+1}^-}^T \mathbf{X}_{n_{t+1}^-}}^{O(m_t^2 n_{t+1}^-)} \\
- \overbrace{(2\mathbf{X}_{n_t - n_{t+1}^-} - \mathbf{X}_{c_{t+1}}^o - 2\mathbf{1}_{n_t - n_{t+1}^-}\Delta\mathbf{b}_{t+1}^T)^T \mathbf{X}_{c_{t+1}}^o}^{O(m_t^2 c_{t+1})} \tag{23}$$

For the last addend of (23), we can further use of the last identity of (6) to reuse the result of $\mathbf{X}_{t+1}$.

$$2\mathbf{X}_{n_t - n_{t+1}^-} - \mathbf{X}_{c_{t+1}}^o - 2\mathbf{1}_{n_t - n_{t+1}^-}\Delta\mathbf{b}_{t+1}^T \\
= 2\mathbf{X}_{t+1}(1:n_{t+1} - n_{t+1}^+,:) + \mathbf{X}_{c_{t+1}}^o \tag{24}$$

The time complexity for update of correlation matrix $\mathbf{R}_{t+1}$ is clearly $O(m_t^2(n_{t+1}^+ + n_{t+1}^- + c_{t+1}))$ by (14) and (23).

All updates are summarized in Algorithm 1. Combining (6) and (13) the total complexity for correlation matrix update is $O(\min\{m_t n_{t+1}, m_t^2(n_{t+1}^+ + n_{t+1}^- + c_{t+1})\})$, which can be considered as quadratic if $n_{t+1}^+ + n_{t+1}^- + c_{t+1}$ is treated as a small constant. This complexity is theoretically down from straightforward complete re-evaluation by power of 1. In practice, we sometimes could expect even better acceleration, as $\Delta\mathbf{b}_{t+1}^T, \mathbf{X}_{c_{t+1}}^o$ etc. are very often sparse.

**Algorithm 1: Algorithm to Update Correlation Matrix After Window Slide**

**Input:** previous $b_t, \mathbf{X}_t^o, \mathbf{X}_t, \Sigma_t, \mathbf{R}_t$, data of current window
**Output:** $b_{t+1}, \mathbf{X}_{t+1}^o, \mathbf{X}_{t+1}, \Sigma_{t+1}, \mathbf{R}_{t+1}$

Scan the current window, compute $\mathbf{X}_{t+1}^o$ and identify $\mathbf{X}_{n_{t+1}^+}^o, \mathbf{X}_{n_{t+1}^-}^o$ and $\mathbf{X}_{c_{t+1}}^o$.

Update $b_{t+1}$ by equation (5);
Update $\mathbf{X}_{t+1}^o$ by equation (6);
Update $\Sigma_{t+1}$ by equation (13);
Update $\mathbf{R}_t$ by equation (14) and (23);

### C. Lanczos Method

In future discussion, for convenience of mathematical analysis, we assume the correlation matrix is normalized to unit total variance by dividing every element by the number of columns, so that the sun of all eigenvalues is 1, the largest eigenvalue, i.e. the principal weight is between range [0,1], and an eigenvalue larger than 0.5 must be the largest eigenvalue. In implementation, we should instead normalize the computed eigenvalue to the range of [0,1] for better numerical stability.

With updated correlation matrix $\mathbf{R}_{k+1}$, our task is to determine if $\mathbf{R}_{k+1}$ has a large eigenvalue, i.e. the principal weight, is larger than a threshold. If so, we consider there exists potential bot visits in the current time window, and continue to find those hosts that have high correlation with the principal component. Several methods suit for this task, including singular value decomposition, which computes the full PCA in cubic time, or Rayleigh quotient iteration, which finds the exact eigenvalue and eigenvector in cubic time [27].

For our purpose, we discussed in III.A that an accurate evaluation of eigenvalue is not necessary. We will investigate *Lanczos method* [17, 28], a numerical method to approximate the eigenvalues, for its use in our application.

Given any symmetric matrix $\mathbf{R} \in \mathbb{R}^{m \times m}$ (the correlation matrix is symmetric) and any non-zero initial vector $\mathbf{x} \in \mathbb{R}^m$, the *Lanczos iteration* is expressed as the following process,

$$\mathbf{v}_0 = \mathbf{0}, \mathbf{v}_1 = \frac{\mathbf{x}}{\|\mathbf{x}\|} \\
\Rightarrow \begin{cases} \tilde{\mathbf{v}}_j = \mathbf{R}\mathbf{v}_{j-1} - \langle \mathbf{R}\mathbf{v}_{j-1}, \mathbf{v}_{j-1} \rangle \mathbf{v}_{j-1} - \|\tilde{\mathbf{v}}_{j-1}\|\mathbf{v}_{j-2} \\ \mathbf{v}_j = \frac{\tilde{\mathbf{v}}_j}{\|\tilde{\mathbf{v}}_j\|} \end{cases}, j = 2,3,\ldots \tag{25}$$

The iteration is guaranteed to terminate at $j = k_0 + 2$ when it founds $\tilde{\mathbf{v}}_{k_0+2} = \mathbf{0}$, where $k_0$ is the smallest positive integer s.t. $\mathbf{R}^{k_0+1}\mathbf{x} \in \text{span}\{\mathbf{x}, \mathbf{R}\mathbf{x}, \ldots, \mathbf{R}^{k_0}\mathbf{x}\}$, and is guaranteed to be well-defined, i.e. $\|\tilde{\mathbf{v}}_j\| \neq 0$, for $j = 1, \ldots, k_0$. It can be proved that $\mathbf{V}_{k_0+1} = [\mathbf{v}_1, \ldots, \mathbf{v}_{k_0+1}]$ is an orthonormal basis of

span$\{\mathbf{x}, \mathbf{Rx}, ..., \mathbf{R}^{k_0}\mathbf{x}\}$. Let $\langle\rangle$ denote the standard inner product, and let $\mathbf{V}_k = [\mathbf{v}_1, ..., \mathbf{v}_k], 1 \leq k \leq k_0 + 1$, then the Lanczos decomposition gives

$$\mathbf{RV}_k = \mathbf{V}_k \mathbf{T}_k + \tilde{\mathbf{v}}_{k+1} \mathbf{e}_k^T, \forall k = 1, ..., k_0 + 1 \quad (26)$$

where $\mathbf{e}_k \in \mathbb{R}^k$ is a vector with only the $k$th component being 1 and all other components being 0, and

$$\mathbf{T}_k = \begin{pmatrix} \langle \mathbf{Rv}_1, \mathbf{v}_1 \rangle & \|\tilde{\mathbf{v}}_2\| & & \\ \|\tilde{\mathbf{v}}_2\| & \langle \mathbf{Rv}_2, \mathbf{v}_2 \rangle & \ddots & \\ & \ddots & \ddots & \|\tilde{\mathbf{v}}_k\| \\ & & \|\tilde{\mathbf{v}}_k\| & \langle \mathbf{Rv}_k, \mathbf{v}_k \rangle \end{pmatrix} \quad (27)$$

is a symmetric tridiagonal matrix. Since $\tilde{\mathbf{v}}_{k+1}$ is orthogonal to every column of $\mathbf{V}_k$, then (26) is equivalent to

$$\mathbf{V}_k^T \mathbf{R} \mathbf{V}_k = \mathbf{T}_k, \forall k = 1, ..., k_0 + 1 \quad (28)$$

One can choose to use either (26) or (28) for their own convenience. The more crucial point here for our purpose is that the eigenvalues and eigenvectors of $\mathbf{T}_k$ can estimate those of $\mathbf{R}$. The eigenvalues and eigenvectors of a symmetric tridiagonal matrix is a basic task in numerical linear algebra, [16, 21, 29-32]. They take advantage of the special form of symmetric tridiagonal matrices to run faster and more accurate than algorithms for general matrices. The eigenvalue approximation is robust because the error can be shown bounded by the following important theorem,

- **Theorem 1**. For any eigenvalue $\hat{\lambda}$ of $\mathbf{T}_k$, there exists an eigenvalue $\lambda$ of $\mathbf{R}$ s.t $|\lambda - \hat{\lambda}| \leq \|\tilde{\mathbf{v}}_{k+1}\|$; or equivalently if $\mathbf{R}$ has $m$ eigenvalues $\{\lambda_1, ..., \lambda_m\}$, then we have $\min_{1 \leq i \leq m} |\lambda_i - \hat{\lambda}| \leq \|\tilde{\mathbf{v}}_{k+1}\|$. Further, if $\tilde{\boldsymbol{\mu}}$ is an eigenvector of eigenvalue $\hat{\lambda}$ of $\mathbf{T}_k$, then there exists an eigenvalue $\lambda$ of $\mathbf{R}$ s.t $|\lambda - \hat{\lambda}| \leq \frac{\|\tilde{\mathbf{v}}_{k+1}\| |\langle \mathbf{e}_k, \tilde{\boldsymbol{\mu}} \rangle|}{\|\tilde{\boldsymbol{\mu}}\|}$.

The theorem is stated with a short proof sketch in [17]. Since $\mathbf{T}_k \in \mathbb{R}^{k \times k}, \tilde{\boldsymbol{\mu}} \in \mathbb{R}^k$, thus $|\langle \mathbf{e}_k, \tilde{\boldsymbol{\mu}} \rangle|$ is the absolute value of the last component of eigenvector $\tilde{\boldsymbol{\mu}}$, and $\frac{|\langle \mathbf{e}_k, \tilde{\boldsymbol{\mu}} \rangle|}{\|\tilde{\boldsymbol{\mu}}\|} \leq 1$, meaning the second error bound is equal or tighter. Since computation of $\tilde{\boldsymbol{\mu}}$ only takes linear time for a tridiagonal matrix, we will use the second bound. $\frac{|\langle \mathbf{e}_k, \tilde{\boldsymbol{\mu}} \rangle|}{\|\tilde{\boldsymbol{\mu}}\|}$ is usually a small value and it decreases with $k$ [17, 33].

The approximation can also conveniently run in an "online" style due to another useful theorem, which implies the maximum eigenvalue of $\mathbf{T}_k$ is non-decreasing with $k$. Thus, if we are not satisfied with the error bound $d$, we can increase $k$ for better result.

- **Theorem 2**. For any symmetric matrix, the maximum eigenvalues of its leading principal submatrices are always non-decreasing.

Based on above, we propose the following to terminate the iteration early for our application. Suppose we use a threshold $\omega > 0.5$ for principal weight, and let $d = \frac{\|\tilde{\mathbf{v}}_{k+1}\| |\langle \mathbf{e}_k, \tilde{\boldsymbol{\mu}} \rangle|}{\|\tilde{\boldsymbol{\mu}}\|}$, then the practical meaning of Theorem 1 is: if we find a large eigenvalue $\hat{\lambda}$ of $\mathbf{T}_k$, then if $\hat{\lambda} - d \geq \omega$, then the largest eigenvalue of the "normalized" correlation matrix $\mathbf{R}$ must exceed $\omega$, and a warning can be immediately raised; in contrast, if $\hat{\lambda} + d < \omega$ and $\hat{\lambda} - d \geq 0.5$, then the largest eigenvalue of $\mathbf{R}$ must not exceed $\omega$, and we can simply wait for next window slide. For $\hat{\lambda} - d < 0.5$, we setup as threshold $c$, and if $\hat{\lambda} - d$ stays below 0.5 for $c$ rounds, then the largest eigenvalue of $\mathbf{R}$ is not likely to exceed $\omega$, and we can continue to next window slide. This is because the largest eigenvalue of $\mathbf{T}_k$ approaches the largest eigenvalue of $\mathbf{R}$ from below, therefore when the estimated eigenvalue plus the error bound stays below 0.5 for many iterations, it becomes increasingly unlikely for the eigenvalue to exceed a threshold $\omega > 0.5$.

Once the largest eigenvalue $\hat{\lambda}$ of $\mathbf{T}_k$ is estimated, its eigenvector $\tilde{\boldsymbol{\mu}}$ can be found trivially, because we can set $\tilde{\boldsymbol{\mu}}(1)$ to any positive number if $\|\mathbf{v}_2\| \neq \hat{\lambda}$, and $\tilde{\boldsymbol{\mu}}(1) = 0$ otherwise, then other components of $\tilde{\boldsymbol{\mu}}$ can be solved iteratively by $\mathbf{T}_k \tilde{\boldsymbol{\mu}} = \hat{\lambda} \tilde{\boldsymbol{\mu}}$, and $\hat{\boldsymbol{\mu}} = \frac{\mathbf{V}_k \tilde{\boldsymbol{\mu}}}{\|\mathbf{V}_k \tilde{\boldsymbol{\mu}}\|}$ is the estimated principal component of $\mathbf{R}$. The correlation between host $j$ and $\hat{\boldsymbol{\mu}}$ can be computed by $\rho_j = \hat{\boldsymbol{\mu}}(j) \sqrt{\hat{\lambda}}$. We recommend finding the first knee point of the descendingly sorted list of all $\rho_j$, e.g. the point before the first sharp slope, which is inspired by the scree plot of PCA; meanwhile we can disregard hosts with $\rho_j < \omega$ in case the knee point is too low. Hosts satisfying both criteria are considered as potential bots.

Finally, we give some more facts that will be used in the algorithm. First a loose bound of eigenvalues of $\mathbf{T}_k$ are given by $\lambda_l \leq \lambda \leq \lambda_u$ where

$$\lambda_l = \max\{\min\{\langle \mathbf{Rv}_1, \mathbf{v}_1 \rangle - \|\tilde{\mathbf{v}}_2\|, \langle \mathbf{Rv}_i, \mathbf{v}_i \rangle - \|\tilde{\mathbf{v}}_{i+1}\| - \|\tilde{\mathbf{v}}_i\|\}, 0\}$$
$$\lambda_u = \min\{\max\{\langle \mathbf{Rv}_1, \mathbf{v}_1 \rangle + \|\tilde{\mathbf{v}}_2\|, \langle \mathbf{Rv}_i, \mathbf{v}_i \rangle + \|\tilde{\mathbf{v}}_{i+1}\| + \|\tilde{\mathbf{v}}_i\|\}, \text{trace } \mathbf{T}_k\} \quad (29)$$
$$i = 2, ..., k$$

Secondly, the $i$th leading principal minor $p_i(\lambda)$ of $\mathbf{T}_k - \lambda \mathbf{I}$ can be computed as

$$p_i(\lambda) = \begin{cases} 1 & i = 0 \\ \langle \mathbf{Rv}_1, \mathbf{v}_1 \rangle - \lambda & i = 1 \\ (\langle \mathbf{Rv}_i, \mathbf{v}_i \rangle - \lambda) p_{i-1}(\lambda) - \|\tilde{\mathbf{v}}_i\|^2 p_{i-2}(\lambda) & i = 2, ..., k \end{cases} \quad (30)$$

and we use $s_k(\lambda)$ to denote the number of sign changes in $p_0, ..., p_k$. A complete Lanczos method based bot detection algorithm is now as below, based on tridiagonal-matrix eigenvalue estimation algorithm in [16, 21], where we innovate on the termination condition for our particular application using Theorem 1 and Theorem 2.

**Algorithm 2: Lanczos Method Based Bot/Botnet Detection Algorithm**

**Input:** streaming logs, principal weight threshold $\omega > 0.5$,
error tolerance $\epsilon_1, \epsilon_2$ for Lanczos method,
$k_l, k_u$ as the min/max size of $\mathbf{T}_k$ for approximation, $k_s$ as the step,
$c$ as the maximum number of iterations when $\bar{\lambda}$ stays below 0.5.

**Output:** potential bots or botnets

```
while window slide do
    Update correlation matrix R as in Algorithm 1, normalized it to unit total variance;
    Randomly initialize non-zero vector v̂₁ and normalized it as v₁;
    Proceed Lanczos iteration until k = k_l by equation (25) and compute T_k by equation (27);
    while k ≤ k_u && λ_u ≤ ω do
        Compute new inner products and norms of T_k as in equation (27);
        Update eigenvalue upper bound λ_u by equation (29);
        k = k + k_s;
    Compute the eigenvalue lower bound λ_l by equation (29);
    λ̄ = λ_m;
    do
        while |λ_l − λ_u| > ε₁(|λ_l| + |λ_u|) do
            λ = λ − (λ_l+λ_u)/2;
            Compute the numer of sign changes s_k(λ̄) in all leading principal minors of T_k − λ̄I;
            if s_k(λ̄) == k then λ_u = λ̄;
            else λ_l = λ̄;
        Compute approximate eigenvector μ̂ of λ̄ as discussed;
        d = |v_{k+1}ᵀμ̂(k)|/|μ̂| by Theorem 1;
        if λ̄ − d ≥ ω then
            Bots/botnets likely exist, yield warning;
            if d ≤ ε₂ || k == k_u then
                Compute postential bots/botnets by correlation with μ̂ as discussed and yield them;
                continue to next window slide;
            Nest the do-while loop here to find more eigenvalues to detect more than one botnet*;
        else if (λ̄ − d ≥ 0.5 && λ̄ + d < ω) || λ̄ + d < 0.5 for more than c iterations then
            continue to next window slide;
        k = k + k_s;
    while k ≤ k_u && λ_u ≤ ω;
```

The total time complexity of Algorithm 2 for one window slide is $O(k^* m^2)$ where $k^*$ is the value of $k$ when the computation for the current window slide ends, and $m$ is the number of distinct hosts. In practice, the average complexity scales near quadratic as the average number of $k^*$ grows slowly with $m$.

For detection of multiple botnets, the part in marked by ⋆ in Algorithm 2 can be modified to compute more eigenvalues of $\mathbf{T}_k$ by nesting a loop the same as the do-while loop except for the total variance is 1 minus the principal weight, and termination conditions need corresponding adjustment. We omit the detail for space.

## IV. EXPERIMENT & EVALUATION

We simulated both single bots and botnet on an ecommerce web server and collected about four months' log data, totaling 315,688,764 Apache access log entries for our experiments, with 3,075,108 distinct request identifiers (URL to website resources) and distinct 2,519,022 host identifiers (IP addresses or hostname). The human-like visit rate is estimated by average interval between two requests of HTML web pages, which is 39 secs for this dataset.

Some of the hosts in the dataset mark themselves as bots. They are mostly search engines, like google-bots, bing-bots, yahoo-bots. However, they cannot be used as gold standard because they are very well-behaved even though they are bots and hard to detect. Thus, we have to run simulation. The simulation is real on the website server so the generated logs are realistic, and it is guaranteed that it is absolutely harmless for the target server. The simulation is done in four modes to mimic different types of bots.

1) Single request: the simulator randomly choose a link from a fixed list and keeps repeatedly visiting the link for two hours, then the simulator picks the next link.
2) Random list: for every visit, the simulator randomly chooses a link from a list to visit.
3) Fixed list: the simulator visits links of a fixed list in its original order.
4) Focused random walk: the simulator works like a crawler – it downloads a webpage, extracts links of particular pattern from it, and then randomly picks the next link to visit.

The four modes can mimic different types of bots. For example, DOS/DDOS might take the form of any of the first three types; click fraud or statistic skewer are usually the second and the third form, visiting a list of desired target links; price scraper or web crawler could be any of the last three types. We have several further comments: 1) multiple log entries could be generated for a single visit; 2) for all modes, the simulator requests at a Gaussian distributed human-like random rate estimated by the true human visits; 3) only one computer with a distinct IP is used to simulate each mode, therefore the simulation is harmless for a website that handles tens of thousands of customers each day.

The simulated logs for each visit are identified, and then duplicated with new host ids and mixed as desired for various experiment purposes. For example, we can mimic single-bots of high visit rate of each type by duplicating them several times in a time window without changing their host id; we can mimic a massive bot net of human-like visit rate by choosing logs of several different visits, duplicate them many times with distinct host ids, and randomly mixed them with existing logs in several consecutive windows. The simulated logs are used as gold standard for performance analysis.

### A. Performance Analysis

This section presents comparison of accuracy, runtime and sensitivity between PCA and our Lanczos-based algorithm. We run PCA with fixed window and Lanczos with both fixed window (denoted by *Lanczos-B*) and sliding window (denoted by *Lanczos-S*). All experiments are restricted to one thread for fairness. Both methods have researches on parallelism as mentioned in section II.A. For this paper, we focus on experiments with one thread.

For our purpose, we define the *accuracy* as the percentage of known bot visits that can be correctly marked by the algorithm. We compare the Lanczos-method based algorithm with the full PCA on our dataset and measure both running time, accuracy and sensitivity. We experiment on both fixed windows and sliding windows of length from 10 mins to 50 mins. Logs for 2100 single bot visits are placed at random time points with 30 times to 50 times faster visit rate than humans; 100 botnets of 10 to 100 hosts are placed at randomly chosen time points with 1 to 5 times human-like visit rate, and for simplicity, we let bots in a botnet start working simultaneously at the chosen time point, and their visits do not overlap. Each bot/botnet visit lasts from 10 mins to 2hrs. Window slide step is 10% of the window length. For parameters of Algorithm 2, we let $\omega = 0.65$, $\epsilon_1 = 10^{-10}$, $\epsilon_2 = 0.01$, $k_l, k_u, k_s$ are set to 10%, 80% and 1% of the number of hosts in the window, and

$c = 25$. The value of $\epsilon_1$ is given by [16, 21]. The choice of $\omega, \epsilon_2, c$ will be experimented in next section. The results for time and accuracy are shown in Table 1, and it provides clear evidence for the advantage of our algorithm against PCA.

|  |  | Single Bots | Botnets[3] | Run Time[3] |
|---|---|---|---|---|
| 10m | PCA | 70.5% | 52.4% | 12s[1]/213s[2] |
|  | Lanczos-B | 70.1% | 50.3% | **3.6s/20s** |
|  | Lanczos-S | **73.5%** | **63.2%** | 5.5s/44s |
| 20m | PCA | 59.0% | 65.6% | 44s/622s |
|  | Lanczos-B | 58.2% | 64.0% | **12s/81s** |
|  | Lanczos-S | **63.4%** | **78.9%** | 15s/121s |
| 30m | PCA | 53.7% | 79.6% | 89s/1473s |
|  | Lanczos-B | 52.6% | 78.1% | **21s/134s** |
|  | Lanczos-S | **59.8%** | **87.9%** | 23s/184s |
| 40m | PCA | 48.1% | 80.9% | 131s/1959s |
|  | Lanczos-B | 47.5% | 79.1% | **27s/177s** |
|  | Lanczos-S | **54.1%** | **88.2%** | 30s/214s |
| 50m | PCA | 42.1% | 77.8% | 163s/2607s |
|  | Lanczos-B | 41.6% | 76.1% | **32s/218s** |
|  | Lanczos-S | **50.3%** | **85.5%** | 35s/282s |

Table 1 Accuracy & runtime comparison of PCA with fixed time window, Lanczos with fixed time window (Lanczos-B) and Lanczos with sliding time window (Lanczos-S). Results in bold are better. (1) Average runtime of the algorithm on all time windows; (2) maximum runtime; (3) detection of botnets with better accuracy and less running time is the main technical purpose of this research.

Overall, we have several conclusions from Table 1: 1) It is not possible to run PCA with sliding window of length longer than 20m, as its computation time will generally exceed the sliding step; 2) Lanczos-B has competitive performance in comparison to PCA for bot detection; 3) Lanczos has much better running time, and on average it grows almost linearly with the window length; 4) Lanczos with sliding window consistently has higher accuracy than fixed window.

A subtler implication of above results is about *choice of window length*. For single bots, longer window length damages performance, which is the weakness of all correlation-based botnet detection algorithms. This is because if there exists more than one "not-so-correlated" single bots in a time window, the principal weight will plunge, and longer time window length increases the probability of this situation. For botnet, within a range, accuracy increases with window length, because more data provide more statistical evidence of correlation if the botnet exists, especially when botnets include randomness in their visit pattern, like the random walk and random list in our simulation. However, longer time window usually makes the algorithm less capable of detecting bots with shorter visit duration. For example, for 50-min window length, botnet visits of less than 20 minutes become less discoverable. Given the assumption that botnets often have to maintain at least a request per 30 to 40 secs (the human-like visit rate of our dataset) to achieve its goal at a reasonable cost, 40 minutes' data generally are sufficient for exposing them. In practice, we would recommend running at least two threads to monitor a log stream, one with short length (like 10-min, 15-min) for discovery of single bots and short-duration visits of botnets, and the other with medium length (like 30-min, 40-min) for detection of other botnets. If detection of slower botnet is desired, we can add one more thread with longer window length.

We now present the *sensitivity* of both PCA and Lanczos-S which is defined as how much time does the algorithm needs to detect the bots since their initial requests. The experiments are done for 10-min window length and 40-min window length to test how what percentage of bots are detected within a specified sensitivity. The results are shown in Figure 3 with a point $(x, y)$ on a curve for algorithm $z$ means $x\%$ of the bots detected by $z$ are detected within $y$ minutes of their initial request. A higher curve implies better sensitivity. The overall average sensitivity results are shown in Table 2. We can observe a clear advantage of Lanczos-S from both Figure 3 and Table 2. The advantage comes from two sources: less computation time of Lanczos, and the sliding window. Ultimately, it is all due to the less time complexity of Lanczos, which enables us to use sliding window rather than fixed windows.

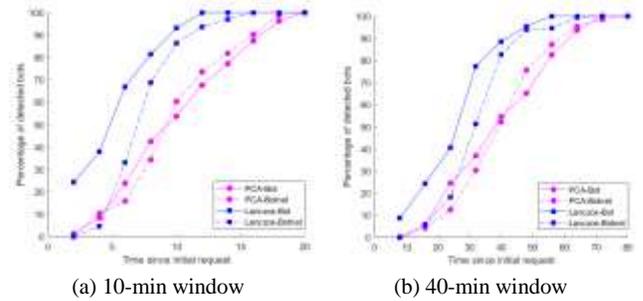

(a) 10-min window  (b) 40-min window

Figure 3 Experiment results for the sensitivity of PCA and Lanczos-S. "-Bot" suffix means results for single bots, "-Botnet" suffix means results for botnet-bots. x-axis marks the time in minutes since initial request, y-axis indicates among bots detected by the algorithm, what percentage is detected within corresponding minutes since initial request. For example, a point at $(x, y)$ on the curve for PCA-Botnet means $x\%$ of all detected botnet-bots are detected within $y$ minutes since their initial request.

|  | 10m-Bot | 10m-Botnet | 40m-Bot | 40m-Botnet |
|---|---|---|---|---|
| PCA | 9.85m | 10.69m | 38.20m | 42.58m |
| Lanczos-S | 4.94m | 7.34m | 25.25m | 33.35m |

Table 2 Results of average sensitivity (average time the algorithm needs to detect the bots since their initial requests) for PCA and Lanczos-S with 10-min and 40-min window lengths.

### B. Experiment on Parameters

This section presents experiments on parameters $\omega, c$ and $\epsilon_2$ and provides some insight into choice of their values. The experiments is on the whole dataset with Lanczos-B, because Lanczos-S takes too much time, and the results of Lanczos-B should be good enough for our discussion. In the case of real botnet detection, we can run the following experiments on historical logs and decide good values for the parameters.

For choice of $\omega$, we first run Lanczos-B on the original data (without simulation) over all fixed 40-min windows, and then use the Markov chain behavior model in section II.B to recognize and remove strangely-behaved suspicious hosts, and then run the Lanczos-B over all windows again. The principal weight distribution change is shown in Figure 4. The histogram shows most suspicious hosts are in window with principal weights higher than 0.5 quite possibly contain bots. However, 1) our algorithm requires that $\omega > 0.5$; 2) we emphasize botnet detection, and mile principal weight is less likely to imply existence of botnet; 3) the rate of false alarm should be contained at a reasonable level. Considering all these factors, we choose $\omega = 0.65$ as our threshold.

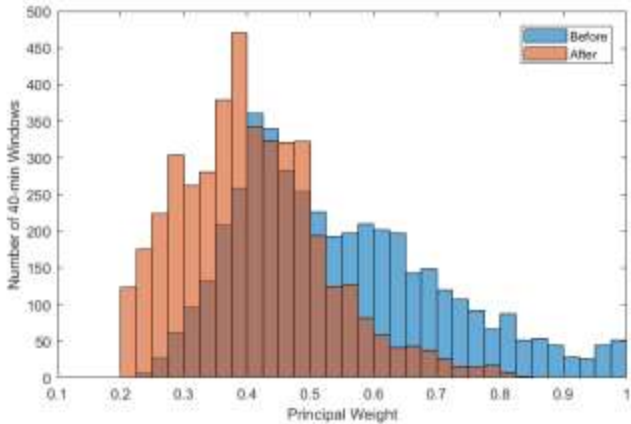

Figure 4 Principal weight distribution change after using the Markov chain behavior model to remove strangely-behaved hosts from the logs of all 40-min fixed time windows.

For choice of $c$, it affects the early termination for a window where bots are not likely to exit, and a smaller $c$ will decease runtime, but possibly introduce certain error. For choice of $\epsilon_2$, it affects the early termination for a window where bots are detected, and smaller $\epsilon_2$ indicates a more accurate principal component is desired. The results using Lanczos-B with both 10-min window and 40-min window are shown in Figure 5, and for the experiments in previous section we choose parameter value at the "elbow" points in Figure 5 (b),(d). We in addition remark that: 1) for shorter time window, larger $c$ may take larger value without hurting much accuracy, thus if we run a second thread with short time window as suggested in choice of window length in previous section, we could specify a larger $c$; 2) $\epsilon_2$ has stronger effect on performance, thus we recommend setting it to a small value even for short time window.

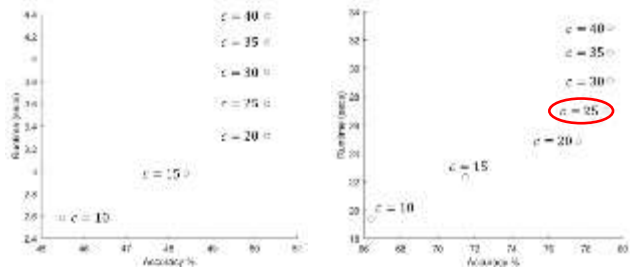

(a) $c$, 10-min window  (b) $c$, 40-min window

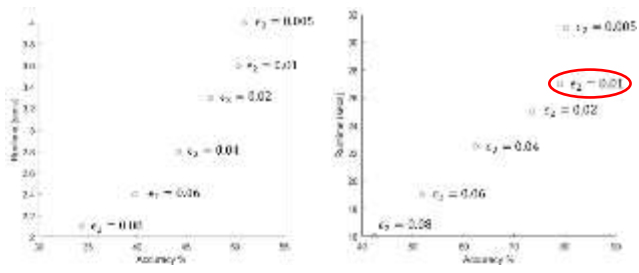

(c) $\epsilon_2$, 10-min window  (d) $\epsilon_2$, 40-min window

Figure 5 Relation between $c, \epsilon_2$ with average runtime and accuracy for Lanczos method. Smaller $c, \epsilon_2$ improves accuracy at cost of more runtime. For experiments in previous section, we choose $c, \epsilon_2$ at the elbow point $c = 25$ and $\epsilon_2 = 0.01$ of (c) and (d).

### C. Botnet Examples

We apply our algorithm on the original data without simulated logs, and 4,557 distinct hosts are marked as potential bots, with 3,417 of them recognized as botnet-bots in 232 potential botnets. Botnets discovered in different time windows are merged if they share two or more hosts. We use an automatic program to compare the IPs with the Barracuda IP reputation database and 88% of them have poor reputation. Some of the botnets are search engine like "crawl-66-249-xx.xxx.googlebot.com" that can be actually confirmed. Besides that, we manually confirmed one of the top-rated botnet that do not mark themselves as bots in the access log, which comes from a website monitor company Anturis, as shown in Figure 6. Some other botnets are demonstrated in Figure 7, which clearly shows that one bot in the botnet might play the role of a controller.

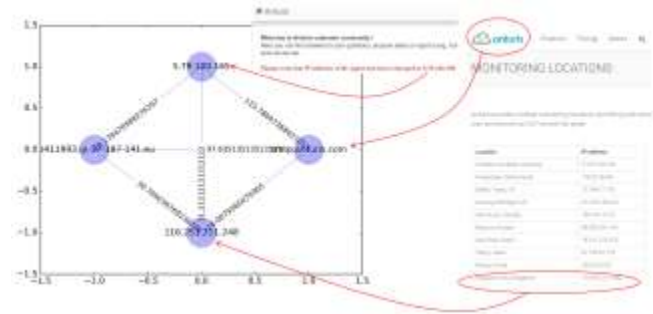

Figure 6 A top-rated botnet is manually confirmed from a company providing website monitoring service. At the time our experiment, they do not mark themselves as bots in the Apache access log.

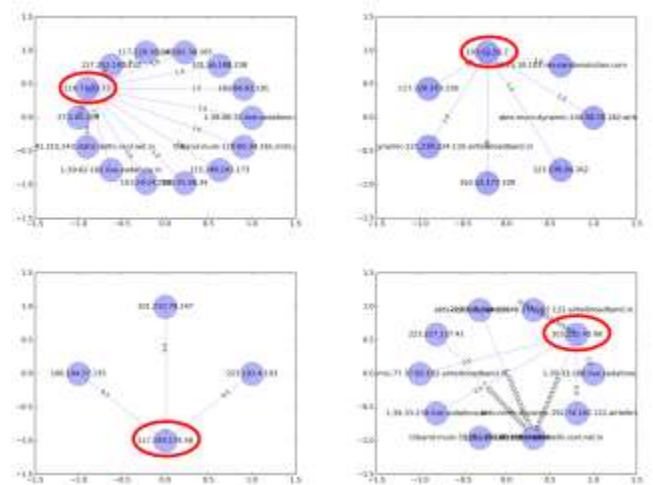

Figure 7 Some other examples of recognized botnets, where we can see usually one bot might play the role of a controller.

## V. CONCLUSION & FUTURE WORK

The main objective of this paper is to recognize botnets from streaming web server logs. We recognize and adapt Lanczos method to the application of botnet detection. For this purpose, we first develop the online correlation matrix updates, and feed them to the Lanczos iterations. Making use of Lanczos error bound the non-decreasing eigenvalues of symmetric matrices, and the special properties of our application, a method is proposed to terminate the iterations early. Our approach improves time use of eigenvalue-based botnet detection from cubic to sub-cubic, which enables us to monitor the log stream by sliding windows, rather than batch-based detection. Experiments show the time cost of Lanczos method with different time windows are consistently only 20% to 25% of PCA. In the future, we could further the research in two directions: 1) finding its good use in other anomaly detection applications; 2) compensate its weakness in single bot detection by the Markov-chain behavior model as mentioned in section II.B.